# Approaching Bio Cellular Classification for Malaria Infected Cells Using Machine Learning and then Deep Learning to compare & analyze K-Nearest Neighbours and Deep CNNs

Dhron Joshi
Computer Science
University of Waterloo
d8joshi@uwaterloo.ca

Rishabh Malhotra
Computer Science
University of Waterloo
r22malho@uwaterloo.ca

Ku Young Shin
Computer Science
University of Waterloo
ky2shin@uwaterloo.ca

*Abstract*—**Malaria is a deadly disease which claims the lives of hundreds of thousands of people every year. Computational methods have been proven to be useful in the medical industry by providing effective means of classification of diagnostic imaging and disease identification. This paper examines different machine learning methods in the context of classifying the presence of malaria in cell images. Numerous machine learning methods can be applied to the same problem; the question of whether one machine learning method is better-suited to a problem relies heavily on the problem itself and the implementation of a model. In particular, convolutional neural networks and k-nearest neighbours are both analyzed and contrasted in regards to their application to classifying the presence of malaria and each model's empirical performance. Here, we implement two models of classification; a convolutional neural network, and the k-nearest neighbours algorithm. These two algorithms are compared based on validation accuracy. For our implementation, CNN (95%) performed 25% better than kNN (75%).**

## I. Introduction

Machine learning is involved in numerous application domains ever since its inception and the ability to process large amounts of data; the use of GPUs has made it even more accessible to process image data using convolutional neural networks (CNNs) or classify data using the k-Nearest Neighbours (kNN) algorithm. Malaria, while not prominent in North America, is a disease that has been rampant throughout time and claiming an estimated 435,000 lives in 2017. Malaria, while detectable by symptoms, is diagnosed confidently via a medical practitioner examining a blood sample. The ability to detect the presence of malaria computationally has many implications for faster medical diagnostics and further treatment. The dataset for which the models were trained on contains 27, 558 images of segmented cells from thin blood smear slide images provided by the National Library of Medicine (LNM). The cells are divided into equal instances of uninfected and parasitized groups taken from 50 uninfected patients and 150 infected patients in Bangladesh.

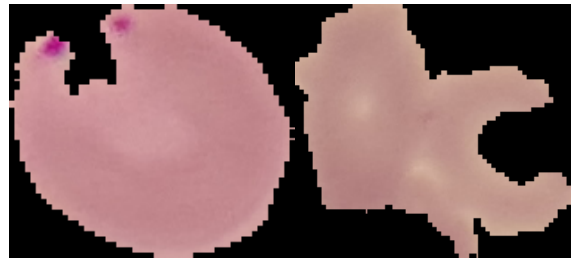

Fig. 1. Parasitized cell on the left, uninfected cell on the right

Implementations of the CNN and kNN algorithm have been used to identify empirical performance results, and we find in general, CNN performs better on the malaria dataset based on accuracy of prediction.

## II. Techniques To Tackle The Problem

We are considering 2 approaches to this complex bio-cellular classification problem. The first is using *Machine Learning: K-Nearest Neighbours(kNN)* and the second being, using *Deep Learning: Deep Convolutional Neural Networks(Deep CNNs)*.

So naturally, the first question we must answer is the basic idea: What should we use for this problem, machine learning or deep learning and to answer that, we must first understand the core differences between machine learning and deep learning.

### A. Deep Learning vs Machine Learning

**Machine Learning** is a subset of the field of artificial intelligence involved with the creation of algorithms which can modify itself without human intervention to produce desired output- by feeding itself through structured data.

On the other hand, **Deep Learning** is a subset of machine learning where algorithms are created and function similar to those in machine learning, but there are numerous layers of these algorithms- each providing a different interpretation to the data it feeds on. Such a network of algorithms are called artificial neural networks, being named so as their functioning is an inspiration, or you may say; an attempt at imitating the function of the human neural networks present in the brain.

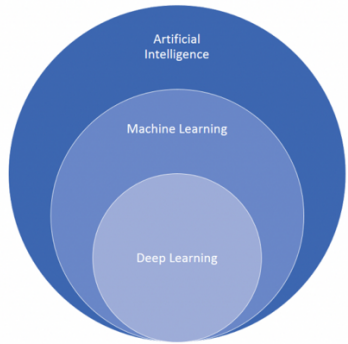

Fig. 2. AI, Machine Learning, and Deep Learning

The basic differentiating factor is that machine learning approaches require well structured and well labelled data (also referred to as training data). For example, in the popular cat-dog classification problem, one may simply label the pictures of dogs and cats in a way which will define specific features of both the animals. This data will be enough for the machine learning algorithm to learn based on which it may continue to classify an infinite amount of additional unseen images as a cat or a dog.

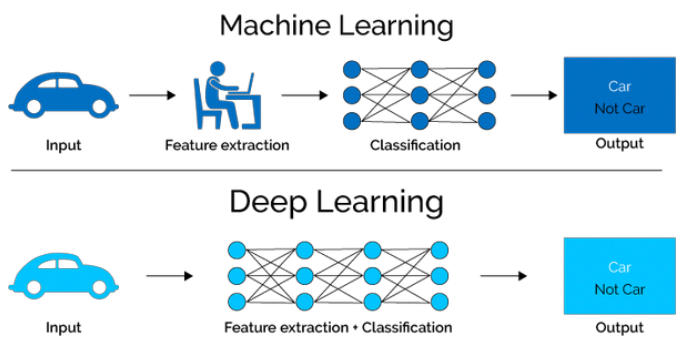

Fig. 3. Machine Learning vs Deep Learning

Deep Learning, however, would take a very different approach to this same cat-dog classification problem. The main advantage of deep learning networks is that they do not necessarily need well structured data(still need everything to be labelled generally speaking, approaches like Generative models & Adversarial approaches or Active Learning tackle this issue of limited labelled data, however, analysis of these techniques is well beyond the scope of this document) of the pictures to classify the two animals. The artificial neural networks which fall under the umbrella of deep learning, send the input (the training data of images) through different layers of the network, with each layer within the network hierarchically defining specific features of images. This functionality that ANNs try to mimic is inspired by the workings of the human brain. After the data is processed through layers within deep neural networks, the system finds the appropriate identifiers for classifying both animals from their images. Feature Selection doesn't have to be done manually with deep learning based approaches which can be a huge plus, especially for problems where identifying and defining features might be next to possible by standards of limited human perception, just like might be the case for a complex problem like cellular imaging which we have. These features, for example might be good or bad differentiating factors and may be perceptible to the human eye after applying certain image processing filters to these images. Which filters may those be? What if certain filters work well with some images and others work well with others? What if certain "good" features are observable only after applying a specific sequence of filters like greyscale + noise reduction? For these reasons, when feature selection becomes next to impossible by standards of human perception and the amount of data available is huge, deep learning might provide amazing performance compared to machine learning. But yes, the disadvantage of deep learning being the amount of data it needs. Deep learning usually might need an exponential amount of data to solve the same that machine learning would. The reason for this being that it is only able to identify edges (concepts, differences) within layers of neural networks when exposed to over a million data points. A naive explanation for this would be that while machine learning knows the gist of features on the basis of which to classify whereas deep learning needs to try out a huge number of features to identify which ones work well for a given dataset. And it is for this reason, deep learning needs monumentally more data than machine learning. Here is a performance analysis comparison of deep learning and machine learning with the amount of data available:

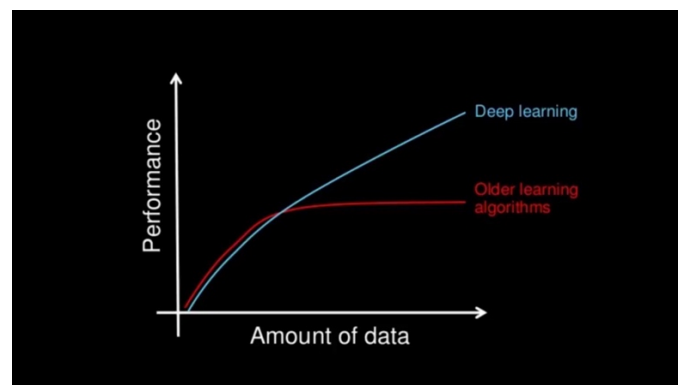

Fig. 4. Deep Learning vs Older learning algorithms

The red curve depicts machine learning performance and the blue one depicts deep learning performance. So, in the event that the following 3 conditions are satisfied, one can depend upon deep learning to deliver better results than classic machine learning algorithms:

• Huge amount of data is available. (Can be circumvented to some extent using techniques like Transfer Learning, which are again out of scope of this document)

• There are ample compute resources(GPUs etc; especially for image based datasets).

• And the problem is complex and large in scale enough that machine learning is unviable.

Now let's take a look at the 2 algorithms which we are using here: kNN and Deep-CNN. K-Nearest Neighbours

*B. K-Nearest Neighbours*

At the most basic level, kNN makes predictions using the training dataset directly. The training data is the model. Predictions are made for a new instance (x) by searching through the entire training set for the K most similar instances (the neighbours) and summarizing the output variable for those K instances.

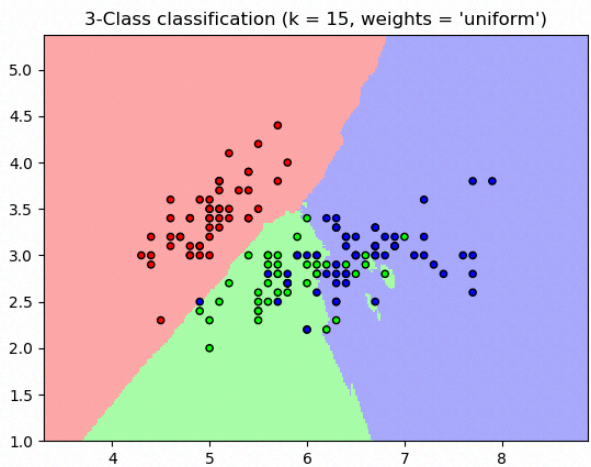

Fig. 5. kNN Classification

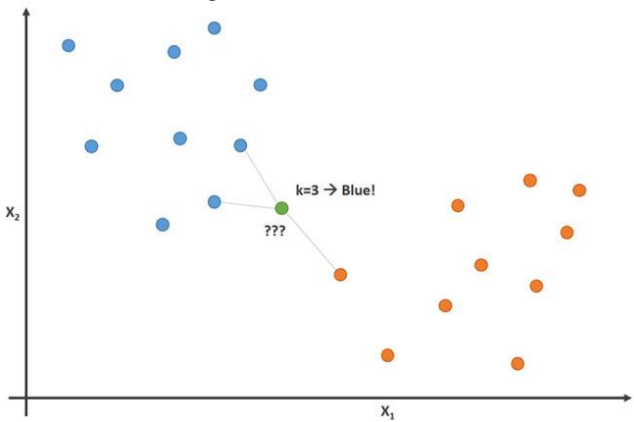

Fig. 6. kNN Decision Function

For regression this might be the mean output variable, in classification this might be the mode (or most common) class value. To determine which of the K instances in the training dataset are most similar to a new input a distance measure is used. There are 4 popular types distances that are usually used in practice:

• **Euclidian Distance**: Euclidean distance is calculated as the square root of the sum of the squared differences between a new point (x) and an existing point (xi) across all input attributes j:

$$EuclideanDistance(p, q) = \sqrt{\sum_{i=1}^{n}(q_i - p_i)^2}$$

• **Hamming Distance**: Calculate the distance between binary vectors.

• **Manhattan Distance**: Calculate the distance between real vectors using the sum of their absolute difference. Also called City Block Distance.

• **Minkowski Distance**: This distance metric is a generalization of Euclidean and Manhattan distance.

The above mentioned list is in no way an inclusive list though and there are however, many other kinds of distance metrics that could be used, for example: Tanimoto Distance, Mahalanobis Distance, Cosine Distance, etc.

When it comes to classification, kNN is used for classification, the output can be calculated as the class with the highest frequency from the K-most similar instances. Each instance in essence votes for their class and the class with the most votes is taken as the prediction.

Class probabilities can be calculated as the normalized frequency of samples that belong to each class in the set of K most similar instances for a new data instance. For example, in a binary classification problem (class is 0 or 1):

$$P(Class = 0) = \frac{Count(Class=0)}{Count(Class=0)+Count(Class=1)}$$

However, all is not well with kNN and it has some significant downsides; the most popular one being what is commonly referred to as The Curse of Dimensionality. kNN works well with a small number of input variables (p), but struggles when the number of inputs is very large. Each input variable can be considered a dimension of a p-dimensional input space. For example, if you had two input variables x1 and x2, the input space would be 2-dimensional. As the number of dimensions increases the volume of the input space increases at an exponential rate. In high dimensions, points that may be similar may have very large distances. All points will be far away from each other and our intuition for distances in simple 2 and 3- dimensional spaces breaks down.

## C. Convolutional Neural Networks

A Convolutional Neural Network (ConvNet/CNN) is a Deep Learning algorithm which can take in an input image, assign importance (learnable weights and biases) to various aspects/objects in the image and be able to differentiate one from the other. The pre-processing required in a ConvNet is much lower as compared to other classification algorithms. While in primitive methods filters are hand-engineered, with enough training, ConvNets have the ability to learn these filters/characteristics.

The architecture of a ConvNet is analogous to that of the connectivity pattern of neurons in the Human Brain and was inspired by the organization of the Visual Cortex. Individual neurons respond to stimuli only in a restricted region of the visual field known as the Receptive Field. A collection of such fields overlap to cover the entire visual area.

A ConvNet is able to successfully capture the Spatial and Temporal dependencies in an image through the application of relevant filters. The architecture performs a better fitting to the image dataset due to the reduction in the number of parameters involved and reusability of weights. In other words, the network can be trained to understand the sophistication of the image better.

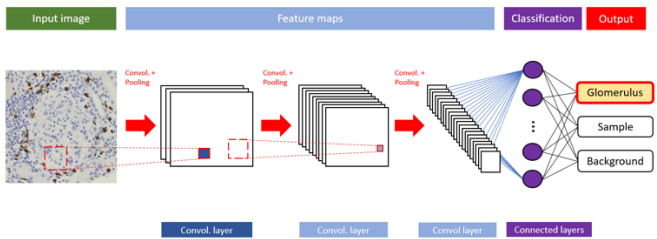

Fig. 7. Convolutional Neural Networks

Now is a good time to see how a CNN works given an input image, let us use a simple RGB image for our understanding.

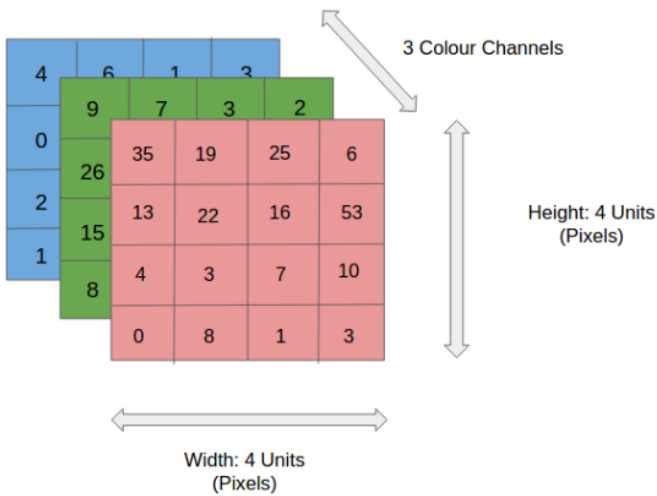

Fig. 8. RGB Image Represented in Three Colour Planes

In Fig. 7, we have an RGB image which has been separated by its three colour planes — Red, Green, and Blue. However, it doesn't have to be an RGB colour space. There are a number of such other colour spaces in which images may exist — Grayscale, HSV, CMYK, etc.

The role of the ConvNet is to reduce the images into a form which is easier to process, without losing features which are critical for getting a good prediction. Of course, this comes in handy when dealing with huge dimensions and dimension spaces can get really large, in fact imagine how computationally intense an image with 8K resolution(7680x4320) could become for a convolutional neural network.

But, neural network is within our realm of understanding for now. What is a convolution? The objective of the Convolution Operation is to extract the high-level features such as edges, from the input image. ConvNets need not be limited to only one Convolutional Layer. Conventionally, the first ConvLayer is responsible for capturing the Low-Level features such as edges, colour, gradient orientation, etc. With added layers, the architecture adapts to the High-Level features as well, giving us a network which has the wholesome understanding of images in the dataset, similar to how we would.

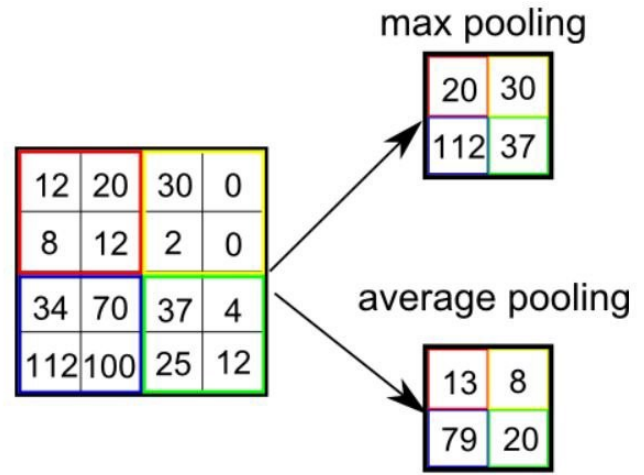

Fig. 9. Max Pooling and Average Pooling

Once a convolution has been completed, the Pooling layer is responsible for reducing the spatial size of the Convolved Feature. This is to decrease the computational power required to process the data through dimensionality reduction. Furthermore, it is useful for extracting dominant features which are rotational and positional invariant, thus maintaining the process of effectively training of the model. There are 2 popular pooling operations: max or average pooling. Max Pooling returns the maximum value from the portion of the image covered by the Kernel. On the other hand, Average Pooling returns the average of all the values from the portion of the image covered by the Kernel.

The Convolutional Layer and the Pooling Layer, together form the i-th layer of a Convolutional Neural Network. Depending on the complexities in the images, the number of such layers may be increased for capturing low-levels details even further, but at the cost of more computational power.

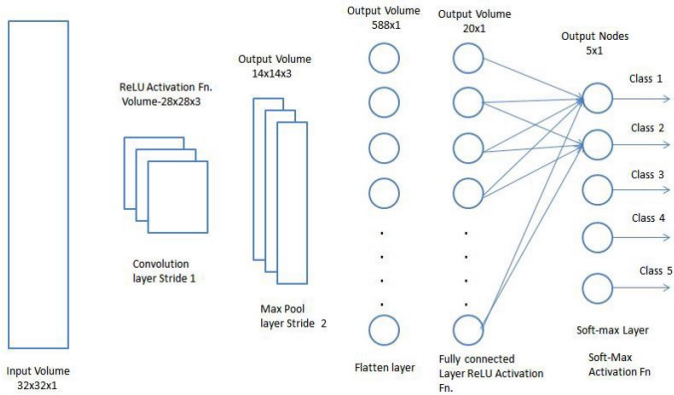

Fig. 10. Neural Network with Convolutional Layer

Finally, we reach to the classification stage which is done using the Fully Connected Layer (FC Layer).

Adding a Fully-Connected layer is a (usually) cheap way of learning non-linear combinations of the high-level features as represented by the output of the convolutional layer. The Fully-Connected layer is learning a possibly non-linear function in that space. Finally, a classification function such as softmax function can be used.

There are various architectures of CNNs available which have been key in building algorithms which power and shall power AI as a whole in the foreseeable future such as: CapsNet, LeNet, AlexNet, VGGNet, GoogLeNet, ResNet, ZFNet and many others.

### III. EMPIRICAL EVALUATION

#### A. Setup

To compare the performances of the mentioned algorithms, kNN and CNN were implemented for this problem.

#### B. kNN

All experiments were done using Python, preprocessing images using OpenCV for kNN. Since kNN classification uses a distance metric to find the k closest neighbours for its predictions, the images needed to be translated into vectors for kNN. To put our training data in vector space, images were resized to a fixed size of 32x32 pixels and then flattened. As well as flattened images in raw pixel values, colours were extracted to a 3D colour histogram from the HSV colour space. Using the processed vectors, predictions were made using Euclidean Distance formula and the models were evaluated using 4-fold cross-validation. To find the most optimal k-value, values from 1 to square root of the dataset ~150 were tested.

#### C. CNN

As for our CNN model, Tensorflow and Keras were used and the neural network was implemented as follows:

```
Layer (type)                 Output Shape              Param #
=================================================================
input_1 (InputLayer)         (None, 64, 64, 3)         0
conv2d (Conv2D)              (None, 31, 31, 64)        1792
batch_normalization_v1 (Batc (None, 31, 31, 64)        256
conv2d_1 (Conv2D)            (None, 15, 15, 128)       73856
dropout (Dropout)            (None, 15, 15, 128)       0
conv2d_2 (Conv2D)            (None, 13, 13, 256)       295168
max_pooling2d (MaxPooling2D) (None, 6, 6, 256)         0
conv2d_3 (Conv2D)            (None, 4, 4, 1024)        2360320
dropout_1 (Dropout)          (None, 4, 4, 1024)        0
conv2d_4 (Conv2D)            (None, 2, 2, 512)         4719104
dropout_2 (Dropout)          (None, 2, 2, 512)         0
flatten (Flatten)            (None, 2048)              0
dense (Dense)                (None, 256)               524544
dropout_3 (Dropout)          (None, 256)               0
dense_1 (Dense)              (None, 2)                 514
=================================================================
Total params: 7,975,554
Trainable params: 7,975,426
Non-trainable params: 128
```

Fig. 11. CNN Used for this Dataset

#### D. Results

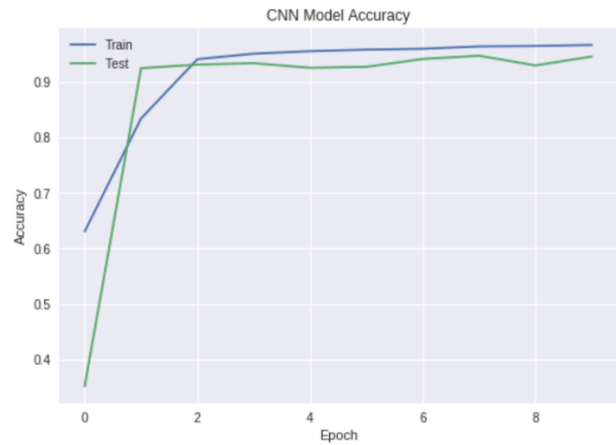

Fig. 12. CNN Model Accuracy

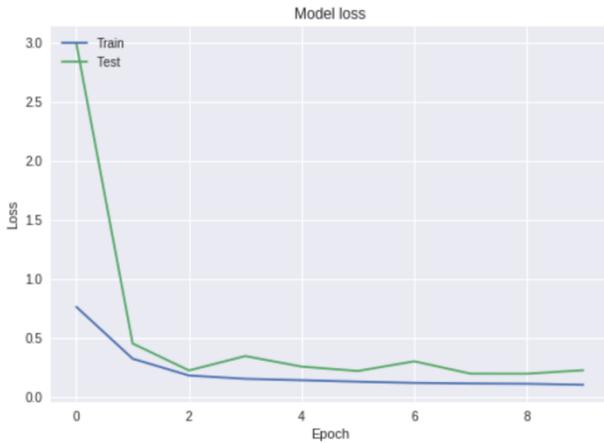

Fig. 13. CNN Model Loss

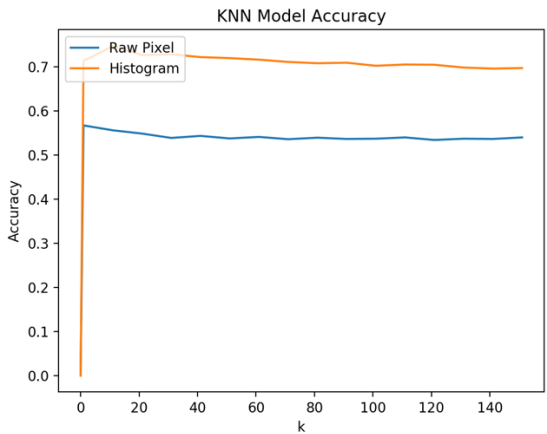

Fig. 14. kNN Model Accuracy

As illustrated in Fig. 14, raw pixel has a validation accuracy of .5564 (55.64%), whereas colour histogram has a validation accuracy of .747 (74.7%) with k-value of 10. The significant difference in raw pixel accuracy and colour histogram accuracy can be described by our dataset. The parasitized cells carry more meaningful information in colour than patterns. By using raw pixel intensities to calculate our euclidean distances, the test data does not get much of a distinction between a parasitized cell and an uninfected cell due to their round shapes. Thus, the results for raw pixel intensities are only slightly better than random guessing (50%).

In the Fig. 12., the CNN model achieved an accuracy of 0.9456 (94.56%). Fig. 13. shows consistent losses over the last 3 epochs, meaning the model did not display any signs of overfitting. As a result, CNN outperformed the kNN model. Since the convolutional layers are much better at feature extraction as to simple colour histogram conversion, the neural network is able to update its kernels more effectively and produce a finer output.

*E. Model Comparison*

The pros of using kNN for this problem is execution time and its complexity. This model is easier to implement since the prediction is finding the class with most occurrence within k nearest neighbours. Because this algorithm calculates the distances for each test data, testing is slow compared to other models like CNN. It is also difficult to find a good distance function for kNN.

A notable pro for CNN is its accuracy in this problem. It is also more robust to noisy data compared to kNN. It can automatically extract the features from a given image, and modify the values in its kernel while learning from training data. However, CNN is computation heavy. It requires powerful and expensive hardware to train the model in a reasonable amount of time, and large amount of training data is needed.

IV. CONCLUSION

In comparison of the empirical performance of kNN and a CNN based on particular their implementations, the CNN implementation outperforms kNN in terms of model accuracy. CNN is particularly more performant at classifying malaria due to its robustness and ability to extract features from the given cell images while the kNN is computationally more feasible with a trade-off of lesser accuracy in classification. The malaria dataset, while effectively classified, now requires more research in the ramifications and implementations of these computational methods in a clinical setting. We recommend research on how implementations like the CNN or kNN classifier can be accessible on a large-scale for performant medical diagnostics and treatment, especially in developing countries.